\documentclass{article}

\usepackage{xcolor}

\usepackage[final, nonatbib]{neurips_2022_ml4ps}
\usepackage{graphicx}
\usepackage[utf8]{inputenc} 
\usepackage[T1]{fontenc}    
\usepackage{url}            
\usepackage{booktabs}       
\usepackage{amsfonts}       
\usepackage{nicefrac}       
\usepackage{microtype}      
\usepackage{xcolor}         

\title{Uncertainty Aware Deep Learning for Particle Accelerators}

\author{%
  Kishansingh Rajput\thanks{kishan@jlab.org
} \\
  Thomas Jefferson National Accelerator Facility,
  Newport News, VA 23606\\
   \And
  Malachi Schram \\
  Thomas Jefferson National Accelerator Facility,
  Newport News, VA 23606\\
  \And
  Karthik Somayaji \\
  University of California Santa Barbara,
  Santa Barbara, CA 93106
}

\begin{document}

\maketitle

\begin{abstract}
  Standard deep learning models for classification and regression applications are ideal for capturing complex system dynamics.
  However, their predictions can be arbitrarily inaccurate when the input samples are not similar to the training data. 
  Implementation of distance aware uncertainty estimation can be used to detect these scenarios and provide a level of confidence associated with their predictions.
  In this paper, we present results from using Deep Gaussian Process Approximation (DGPA) methods for errant beam prediction at Spallation Neutron Source (SNS) accelerator (classification) and we provide an uncertainty aware surrogate model for the Fermi National Accelerator Lab (FNAL) Booster Accelerator Complex (regression).
\end{abstract}

\section{Introduction}
\label{Intro}
Almost all theories in Machine Learning (ML) are based on the assumption that the data used for training a model is Independent and Identically Distributed (IID) with the test/inference data. 
This is not always true because the underlying system can change causing modifications to the data distributions that create samples that are not part of the original training distribution. 
These orthogonal data samples are called out-of-distribution (OOD) samples and the model is not guaranteed to produce accurate predictions on these data.
Widely used ML method, deep neural networks \cite{Carleo:2019ptp, RevModPhys.91.045002, Goodfellow-et-al-2016} are highly expressive due to their large number of parameters and are ideally suited to model complex systems.
Unfortunately, the use of these models for online safety critical applications is limited due to their inability to provide reliable predictions for OOD data samples. 
This problem is exasperated because these models cannot provide uncertainty estimations for their predictions therefore they are unable to inform the application when predictions are unreliable.
For methods that include uncertainty estimations, the predicted uncertainties can be broadly divided into two categories a) model uncertainty and b) data uncertainty. 
In recent years there have been a number of efforts to develop uncertainty quantification methods for deep learning models that include, but are not limited to, Ensemble methods~\cite{https://doi.org/10.48550/arxiv.2007.08792}, Bayesian Neural Networks (BNN)~\cite{https://doi.org/10.48550/arxiv.1506.02142}, and Deep Quantile Regression (DQR)~\cite{koenker_2005}.
Ensemble methods and BNNs both require post-training calibration for the accurate uncertainty estimations and  multiple models or multiple inference calls to estimate the prediction uncertainties.
These shortcomings limit their use for applications that require single inference or when applied on limited memory resources. 
In contrast, DQR is intrinsically calibrated for in-distributed samples, however, it is not explicitly distance aware which limits its ability in estimating OOD uncertainties~\cite{https://doi.org/10.48550/arxiv.2209.07458}.
 Deep Gaussian Process Approximation (DGPA) is an approach that provides a single inference prediction that is calibrated and captures OOD uncertainties by design.
 
In this paper, we present two particle accelerator studies using the DGPA method: 1) errant beam detection at SNS accelerator at Oak Ridge National Laboratory (ORNL), and 2) data driven surrogate model for the FNAL booster accelerator complex.
In Section \ref{methods}, we describe DGPA models for both classification and regression applications and present the results in Section \ref{results}.

\section{Methods}
\label{methods}

In this work, we developed two uncertainty aware deep learning models for both classification and regression using the DGPA method. 
Using this method allows us to leverage the highly expressive nature of deep neural networks as well as the intrinsic distance awareness property of the Gaussian Process (GP)~\cite{Rasmussen2004}. 
A robust uncertainty quantification method is expected to provide uncertainty values that are consistent with the fluctuation of the labeled data and high uncertainty values for the predictions on OOD data samples that can be arbitrarily wrong.

\subsection{Uncertainty Aware Classification Model}
\label{method:classification}
For the classification model, we combine the Spectral-normalized Neural Gaussian Process (SNGP)~\cite{https://doi.org/10.48550/arxiv.2006.10108} technique with a Siamese Neural Network (SNN)~\cite{Koch2015SiameseNN} to predict errant beam pulses at the SNS accelerator before they occur. 
This could potentially allow the shift worker to take preemptive actions.
The SNGP-SNN model is trained on data that represent pulses preceding errant beam pulses.  
Being a similarity based model, SNN can detect unseen anomalies.
The model is trained on Differential Current Monitor (DCM) data and the aim of SNGP-SNN model is to learn the similarities between normal-normal and normal-anomalous beam pulses (here anomalous beam pulses refers to pulses preceding errant beam pulses), and provide a similarity ranking.
By using a similarity ranking metric, it can detect unseen anomalies that were not included in the training data set. 
Additionally, having a robust uncertainty quantification method can indicate whether a detected anomaly is within the training distribution or a new previously unseen anomaly.

A SNN model consists of twin networks accepting unique inputs and providing a similarity score. 
The twin networks architecture is built using four ResNet~\cite{He2016DeepRL} blocks. 
Apart from skip connections, each ResNet block contains 1-dimensional convolution layers with 16, 32, 64, 128 filters respectively with a kernel size of 3, and stride of 2. 
Following each convolution layer is batch-normalization, max-pooling (size 2), and a relu activation layer. 
To avoid over-fitting we added dropout layers (with drop percentage of 5\%) at the end of each ResNet block. 
The twin networks produce a reduced embedding of the inputs capturing salient features.
The embeddings are then compared using a similarity metric defined in Equation~\ref{eq:euclidian}.
\begin{equation}\label{eq:euclidian}
L^{2} = \left| \sum_{i=0}^{N} (x_{1,i}^{2} - x_{2,i}^{2}) \right| 
\label{eq:l2_metric}
\end{equation}
Here $x_{1}$ and $x_{2}$ are the embedding from twin models for input samples 1 and 2, and $i$ is the element wise index. 
The similarity score is then passed to a modified contrastive loss function via a dense layer with 128 nodes and relu activation to maximize the difference between similarity prediction.
The modified contrastive loss is defined in Equation~\ref{eq:closs}.
\begin{equation}\label{eq:closs}
L(y, y') = \alpha \times (1-y) \times y'^{2} + (1-\alpha) \times y \times (max(\beta - y', 0))^{2}
\label{eq:contrastive-loss}
\end{equation}
Where $y$ is the true labels, $y'$ is the predicted similarity score. 
The $\alpha$ and $\beta$ values are used to focus on normal or anomalous samples respectively.

We extended the SNN model to make it distance aware by replacing the output layer with an approximation to GP as described in~\cite{https://doi.org/10.48550/arxiv.2006.10108}. 
A traditional neural network maps the input space to output through hidden layers in the model.
By using a GP approximation, the model can learn to produce uncertainty values that include OOD.
For more details on the SNS system, data preparation and deployment, please refer to \cite{https://doi.org/10.48550/arxiv.2110.12006}. 

\begin{figure}[hb]
  \centering
    \includegraphics[width=\textwidth]{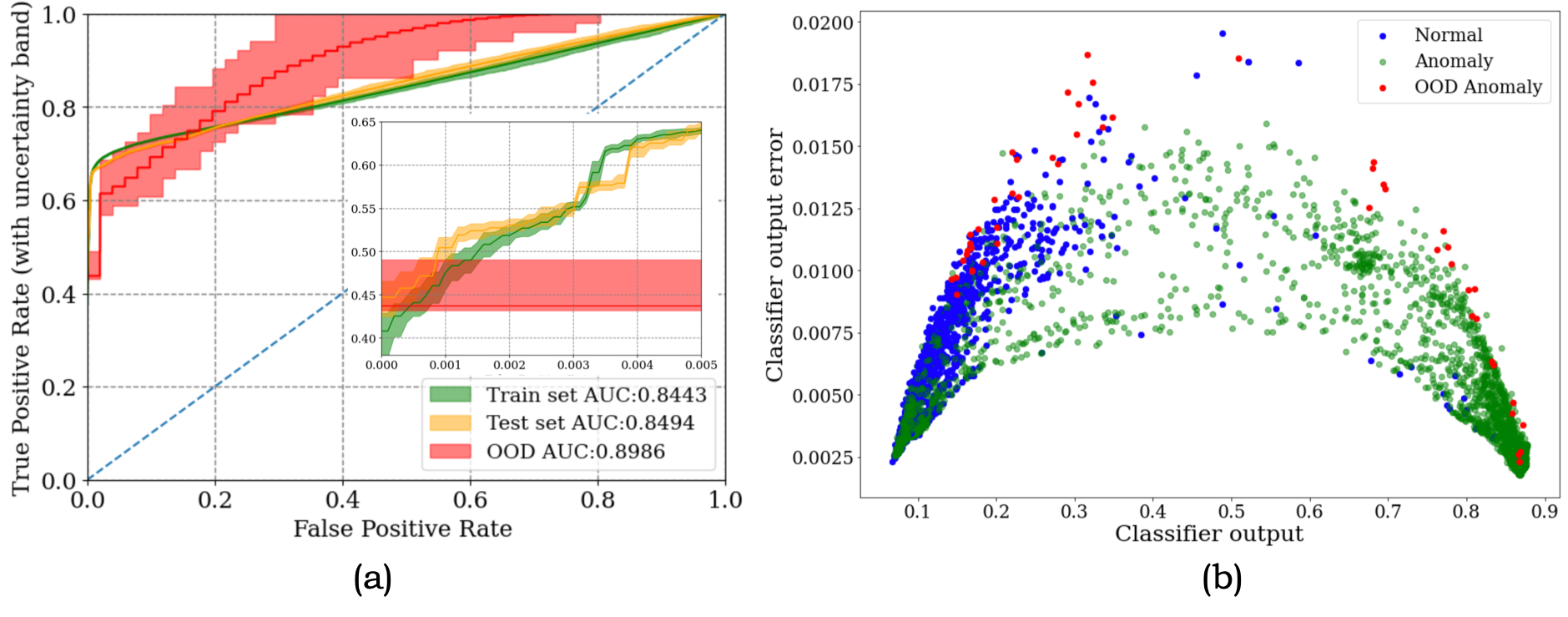}
    \setlength{\belowcaptionskip}{-15pt} 
    \setlength{\abovecaptionskip}{-15pt} 
    \caption{The results on errant beam predictions from SNGP-SNN model. (a) ROC curves with the bands created by smearing the predictions with associated uncertainty values. (b) The scatter plot representing classifier output vs uncertainty values.}
    \label{fig:Siamese}
\end{figure}

\subsection{Uncertainty Aware Regression Model}
\label{method:regression}
For the regression model, we took inspiration from the SNGP paper~\cite{https://doi.org/10.48550/arxiv.2006.10108} and developed an equivalent method for regression problems. 
We used this model to develop an uncertainty aware ML-based surrogate model of the FNAL Booster Accelerator complex for offline optimization problems.
It is critical to include the OOD uncertainty in these applications in order to inform the optimization algorithm when it's no longer in a well modeled area.
We used the historical operational data\cite{BOOSTR:datasheet} with the existing Proportional Integral Derivative (PID) controller to train our surrogate model.
The input data for the surrogate model was reduced to five temporal variables and predicted the next time step for one variable.
The model consists of three 1-dimensional convolution layers with 256 features, kernel size of 3, and a tanh activation, each followed by a batch-normalization, max-pooling (size 2) and a dropout (10\%) layer. 
At the end, the output of the last convolution block is flattened and passed through a dense layer with 256 nodes and tanh activation before the output layer. 
These parameters were chosen after minimal hyper-parameter tuning.

We placed a GP approximation layer as the output layer of the model to provide uncertainty estimation along with the predictions.
We approximated the GP RBF kernel by using 256x256 Random Fourier Features (RFF) matrix.
In order to make the model distance preserving we apply the \textit{bi$-$Lipschitz} constraint, shown in Equation~\ref{eq:biLips}, to a Mean Absolute Percentage Error loss function. 
\begin{equation}\label{eq:biLips}
L_{1} \times ||x_{1} - x_{2}|| \leq ||h_{x_{1}} - h_{x_{2}}|| \leq L_{2} \times ||x_{1} - x_{2}||
\end{equation}
Here $x$ is the input feature vector, $h_{x}$ be the last hidden layer output, and the soft constraint were set to $L_{1}=0.75$ and $L_{2}=1.25$.

\section{Results}
\label{results}
In this section, we describe the results for the errant beam anomaly prediction and the FNAL Booster Accelerator Complex surrogate model. 
The results were produced by using Nvidia Titan-T4 GPUs made available by Thomas Jefferson National Accelerator Facility scientific computing.
Training the models presented in this paper took about 2-3 hours on a Nvidia Titan-T4 GPU. 

\subsection{Classification Model: Errant Beam Prediction at SNS Accelerator}
The SNGP-SNN model was trained using normal beam pulses and one type of anomalous beam pulses (type-0011). 
The SNN sample was created by taking the combinatoric of normal-normal samples (labeled 0) and normal-anomaly samples (labeled 1). 
Once the model was trained it was tested on two different anomaly types 0011 and 1111. 
For more information on data preparation, please refer to the original paper \cite{https://doi.org/10.48550/arxiv.2110.12006}.
The model was able to detect more than $40\%$ of the errant beam pulses for the false positive rate of less than 0.5\% for both anomaly types. 
Figure~\ref{fig:Siamese} (a) shows the ROC curves for the SNGP-SNN model. 
The bands are created with 250 trial predictions sampled from Gaussian distribution with mean equal to the model predicted means and standard deviations equal to the predicted uncertainties.
As can be seen in the Figure~\ref{fig:Siamese}, the uncertainty values for the type 1111 are on average higher than the uncertainty values associated for the predictions for the anomaly type 0011. 
This is because samples of anomaly type 1111 were not added to the training data set and they are seen as OOD by the model.

\subsection{Regression Model: Data Driven FNAL Booster Accelerator Complex Surrogate Model}
The DGPA technique for the FNAL Booster Accelerator Complex surrogate model was built using three 1-dimensional convolutional layers and extended with GP approximation to provide uncertainties to the predictions without any offline calibration.
The raw data was restructured so that each input variable to the model is a time series of 15 timestamps to predict the next time-step forward in the output variable.
The samples were further filtered by explicitly excluding contributions when the main injector lower bound current (I:IB) had a value that exceeded 0.995. 
This filter was used to create in-distribution only training samples that would prevent the model from seeing the cyclic high amplitude in the key input variable (B:VIMIN).   
The data samples with the cyclic high amplitude in B:VIMIN would be considered OOD for the model. For more details on the data preparation please refer to \cite{https://doi.org/10.48550/arxiv.2209.07458}.

\begin{figure}[h]
  \centering
    \includegraphics[width=\textwidth]{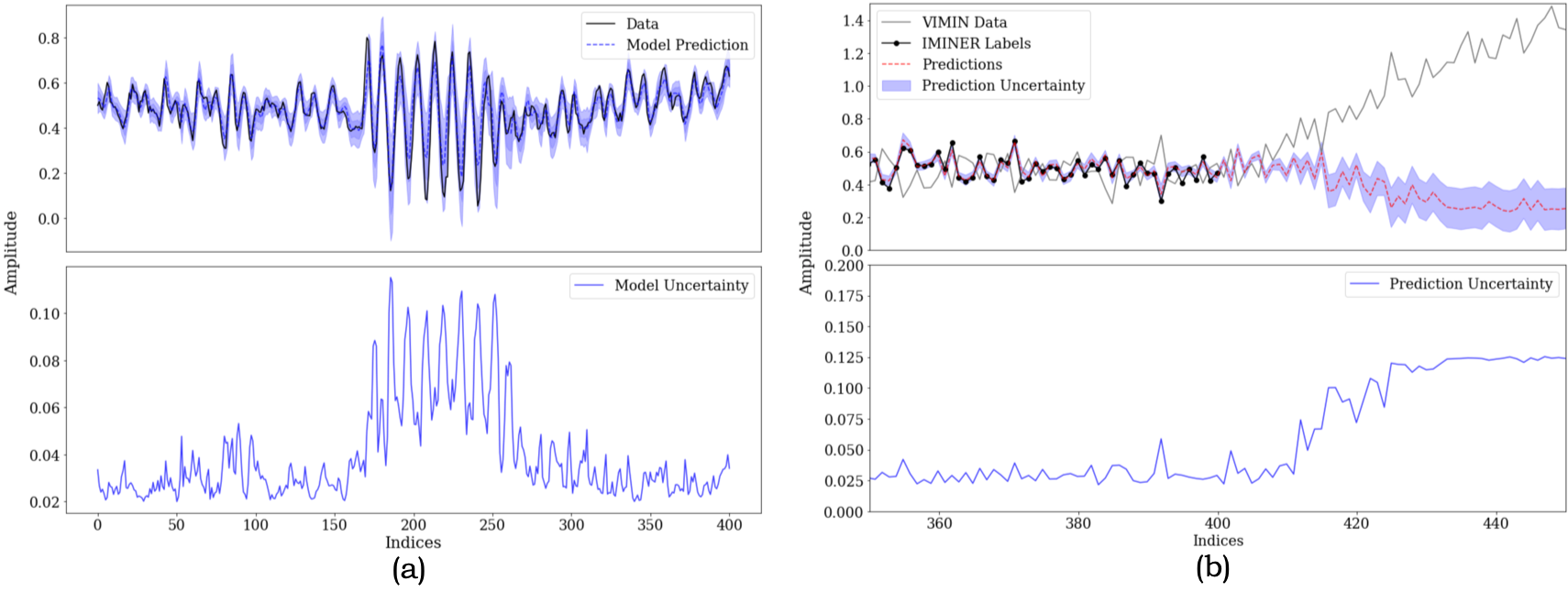}
    \setlength{\belowcaptionskip}{-15pt} 
    \setlength{\abovecaptionskip}{-15pt} 
    \caption{The results from DGPA surrogate model of the FNAL Booster Accelerator. (a) Shows the predictions on the in-distribution and OOD samples along with the associated uncertainty values. The middle region with the high frequency component on the time series represents OOD samples while the initial and tail-end regions represent in-distribution data samples. (b) Shows the predictions and uncertainty values for the synthetic case where the data is intentionally made to enter an OOD region.}
    \label{fig:DGPA}
\end{figure}

Figure~\ref{fig:DGPA} (a) shows the predictions along with the uncertainties.
It can be seen that the model produces higher uncertainties for the cyclic high amplitude region (between timestamps 175-225) where as the in-distribution samples, outside this region, are lower by at least a factor of 2. 
When using this model as an offline environment for Reinforcement Learning (RL) the agent can take different action paths that might lead to OOD states.
The additional uncertainty information provided by the DGPA model can be used by the RL agent to avoid these OOD states.
To illustrate this scenario we monotonically increase a key input variable B:VIMIN until it enters OOD states producing higher uncertainty values, as shown in Figure-\ref{fig:DGPA} (b).

\section{Conclusion}
\label{conclusion}
In this paper, we presented uncertainty aware deep learning models that include GP approximation techniques for both classification and regression applications. 
We presented a new SNGP-SNN model used to detect upcoming errant beam at SNS at the ORNL.
These results supersedes previous published results~\cite{2020-Rescic-AccelFailure,RESCIC2022166064} and provides uncertainty estimations.
We also described how we developed an uncertainty aware surrogate model for the FNAL booster accelerator using a similar GP approximation for regression. 
In the end, we presented the results for both the use-cases. 
The DGPA method discussed is capable of  producing both in-distribution and out-of-distribution uncertainties accurately without offline calibration.
At the same time it is important to note that the kernel approximation in DGPA is limited by the size of the RFF matrix and could limit how well the model can capture the uncertainty.
In the future, we would like to explore the effect of different RBF kernel approximation sizes on the uncertainty estimation.

\newpage
\section{Impact Statement}
We hope the research presented in this paper on data-driven uncertainty aware deep learning methods will provide robust solutions for applications, such as real-time ML applications, anomaly detection, data-driven ML-based surrogate models, and ML-based controls. 
By combining deep learning with uncertainty estimation, that includes OOD, we are able to provide the required information to make a statistically informed decision for complex system. 
The technique presented was explicitly considered to provide a single inference calibrated prediction with uncertainty.
The need for uncertainty aware deep learning models spans a broad spectrum of applications space from particle accelerator control challenges to medical applications. 

\section{Acknowledgements}
This manuscript has been authored by Jefferson Science Associates (JSA) operating the Thomas Jefferson National Accelerator Facility for the U.S. Department of Energy under Contract No. DE-AC05-06OR23177. Part of this research is also supported by  Office of Advanced Scientific Computing Research under Award Number DE-SC0021321. This research used resources at the Spallation Neutron Source, a DOE Office of Science User Facility at Oak Ridge National Laboratory operated by UT Battelle LLC under contract number DE-AC05-00OR22725, and publicly released data sets from Fermi National Accelerator Facility. The US government retains and the publisher, by accepting the article for publication, acknowledges that the US government retains a nonexclusive, paid-up, irrevocable, worldwide license to publish or reproduce the published form of this manuscript, or allow others to do so, for US government purposes. DOE will provide public access to these results of federally sponsored research in accordance with the DOE Public Access Plan (http://energy.gov/downloads/doe-public-access-plan)

\bibliographystyle{unsrt}
\bibliography{main}

\end{document}